\documentclass[11pt]{article}
\usepackage{twocol}
\usepackage{times}
\usepackage{url}
\usepackage{latexsym}
\usepackage{amsmath}
\usepackage{booktabs}
\usepackage{xcolor}
\usepackage{framed,multirow}
\usepackage{graphicx}
\usepackage{subcaption}
\usepackage[normalem]{ulem}
\usepackage{hyperref}
\hypersetup{colorlinks,citecolor=black, linkcolor=black, urlcolor=black}	
\usepackage[nameinlink,capitalize]{cleveref}

\usepackage{authblk}

\usepackage{acro}
\DeclareAcronym{art}{short = ART, long  = approximate randomization testing, cite = Riezler05}
\DeclareAcronym{bleu}{short = BLEU, long  = bilingual evaluation understudy, cite = Papineni02}
\DeclareAcronym{bpe}{short = BPE, long  = byte pair encoding, cite = Gage94}
\DeclareAcronym{cbmt}{short = CBMT, long  = character-based machine translation}
\DeclareAcronym{cbsmt}{short = CBSMT, long  = character-based statistical machine translation}
\DeclareAcronym{cbnmt}{short = CBNMT, long  = character-based neural machine translation}
\DeclareAcronym{cm}{short = CM, long  = confidence measures}
\DeclareAcronym{cer}{short = CER, long  = character error rate}
\DeclareAcronym{fda}{short = FDA, long  = feature decay algorithm, cite = Biccici15}
\DeclareAcronym{hmm}{short = HMM, long  = hidden Markov alignment models, cite = Vogel96}
\DeclareAcronym{htr}{short = HTR, long  = handwritten text recognition}
\DeclareAcronym{il}{short = IL, long  = incremental learning}
\DeclareAcronym{imt}{short = IMT, long  = interactive machine translation}
\DeclareAcronym{inmt}{short = INMT, long  = interactive neural machine translation}
\DeclareAcronym{ismt}{short = ISMT, long  = interactive statistical machine translation}
\DeclareAcronym{ksr}{short = KSR, long  = key stroke rate}
\DeclareAcronym{llm}{short = LLM, long  = large language model}
\DeclareAcronym{lstm}{short = LSTM, long  = long short-term memory, cite = Hochreiter97}
\DeclareAcronym{mar}{short = MAR, long  = mouse action rate}
\DeclareAcronym{mt}{short = MT, long  = machine translation}
\DeclareAcronym{nlp}{short = NLP, long  = natural language processing}
\DeclareAcronym{nmt}{short = NMT, long  = neural machine translation}
\DeclareAcronym{pe}{short = PE, long  = post-editing}
\DeclareAcronym{rbmt}{short = RBMT, long  = rule-based machine translation}
\DeclareAcronym{relu}{short = ReLU, long  = rectified linear unit}
\DeclareAcronym{rnn}{short = RNN, long  = recurrent neural network, cite = {Hochreiter97}}
\DeclareAcronym{sgd}{short = SGD, long  = stochastic gradient descend}
\DeclareAcronym{smt}{short = SMT, long  = statistical machine translation}
\DeclareAcronym{ter}{short = TER, long  = translation error rate, cite = Snover06}
\DeclareAcronym{wer}{short = WER, long  = word error rate}
\DeclareAcronym{wsr}{short = WSR, long  = word stroke rate}
\DeclareAcronym{xml}{short = XML, long  = eXtensible Markup Language}

\newcommand{\x}{\mathbf{x}}
\newcommand{\y}{\mathbf{y}}

\DeclareMathOperator*{\argmax}{arg\,max~}
\title{Two Spelling Normalization Approaches Based on Large Language Models}

\date{}

\begin{document}
\author[1]{Miguel Domingo}
\author[2]{Francisco Casacuberta}

 \affil[1,2]{PRHLT Research Center \protect\\ Universitat Polit\`{e}cnica de Val\`{e}ncia, Spain \protect\\ ValgrAI - Valencian Graduate School and Research Network for Artificial Intelligence, Spain \protect\\ \texttt{\{midobal,fcn\}@prhlt.upv.es}}

\maketitle
\begin{abstract}
    The absence of standardized spelling conventions and the organic evolution of human language present an inherent linguistic challenge within historical documents, a longstanding concern for scholars in the humanities. Addressing this issue, spelling normalization endeavors to align a document's orthography with contemporary standards. In this study, we propose two new approaches based on large language models: one of which has been trained without a supervised training, and a second one which has been trained for machine translation. Our evaluation spans multiple datasets encompassing diverse languages and historical periods, leading us to the conclusion that while both of them yielded encouraging results, statistical machine translation still seems to be the most suitable technology for this task.
\end{abstract}

\section{Introduction}
\label{se:intro}
The orthography of historical texts varies significantly due to the absence of standardized spelling conventions and the evolving nature of human language. As noted by Laing et al. \citet{Laing93}, the Linguistic Atlas of Late Medieval English (LALME) documents a striking diversity, with 45 distinct forms recorded for the pronoun \emph{it}, 64 for \emph{she}, and over 500 for the preposition \emph{through}. Such linguistic variation has long been a central concern for scholars in the humanities.

As historical documents constitute a vital aspect of our cultural heritage, there is a growing interest in their efficient natural language processing \citep{Bollman18}. Nevertheless, the linguistic challenges mentioned earlier pose an additional hurdle. To address these challenges, spelling normalization endeavors to establish orthographic consistency by aligning a document's spelling with modern standards. \cref{fi:Example} illustrates an example of normalizing the orthography of a text.

\begin{figure}[!ht]
	\centering
	%\small
	\begin{minipage}{0.01\textwidth}
		$ $
	\end{minipage}
	\begin{minipage}{0.46\textwidth}
		\underline{\textbf{Original version}}

		{\textquestiondown}C{\'o}mo est\textbf{\textcolor{red}{ay}}s, Ro\textbf{\textcolor{red}{z}}inante, tan delgado?
		
		Porque nunca se come, y se trabaja.
		
		Pues {\textquestiondown}qu{\'e} es de la ce\textbf{\textcolor{red}{u}}ada y de la paja?
		
		No me de\textbf{\textcolor{red}{x}}a mi amo ni \textbf{\textcolor{red}{v}}n bocado.
	\end{minipage}
	\begin{minipage}{0.04\textwidth}
		$ $
	\end{minipage}
	\vspace{10pt}
	\begin{minipage}{0.45\textwidth}
		\vspace{15pt}
		\underline{\textbf{Normilized version}}

		{\textquestiondown}C{\'o}mo est\textbf{\textcolor{teal}{\'ai}}s, Ro\textbf{\textcolor{teal}{c}}inante, tan delgado?
		
		Porque nunca se come, y se trabaja.
		
		Pues {\textquestiondown}qu{\'e} es de la ce\textbf{\textcolor{teal}{b}}ada y de la paja?
		
		No me de\textbf{\textcolor{teal}{j}}a mi amo ni \textbf{\textcolor{teal}{u}}n bocado.
	\end{minipage}
	\caption{Example of adapting a document's spelling to modern standards. Characters that need to be adapted are denoted in \textbf{\textcolor{red}{red}}. Its modern versions are denoted in \textbf{\textcolor{teal}{teal}}. Example extracted from \emph{El Quijote} \protect\citep{Jehle01}.}
	\label{fi:Example}
\end{figure}

Recently, the state of the art of many tasks has shifted towards the usage of pre-trained \ac{llm}, which can be fine-tuned to be adapted to new tasks instead of having to train a system from scratch. Since a frequent problem in the field of historical data is the scarce availability of training resources, in this work we propose two different approaches for spelling normalization based on \ac{llm}s. Our main contributions are as follows:

\begin{itemize}
	\item A new spelling normalization approach based on an \ac{llm} trained without a supervised training.
	\item A new spelling normalization approach based on an \ac{llm} trained for \ac{mt}.
\end{itemize}
\section{Related Work}
\label{se:work}
Various methodologies for spelling normalization have been explored in the literature. Baron et al. \citet{Baron08} proposed the development of an interactive tool equipped with spell checking techniques to aid users in identifying spelling variations. Porta et al. \citet{Porta13} suggested a comprehensive approach involving a weighted finite-state transducer, a modern lexicon, a phonological transcriber and a set of rules. Scherrer et al. \citet{Scherrer13} introduced a method combining historical and modern word lists with character-based \ac{smt}.

Bollman et al. \citet{Bollmann16} implemented a multitask-learning approach utilizing deep bi-\ac{lstm} networks at the character level. Ljubevsic et al. Ljubevsic et al. \citet{Ljubevsic16} applied a token/segment-level character-based \ac{smt} approach to normalize historical and user-created words. Korchagina et al. \citet{Korchagina17} explored \ac{rbmt}, \ac{cbmt} and \ac{cbnmt}.

Domingo and Casacuberta \citet{Domingo18a} evaluated word-based and character-based \ac{mt} approaches, favoring character-based methods, and found that \ac{smt} systems outperformed \ac{nmt} systems. However, Tang et al. \citet{Tang18} compared various neural architectures and concluded that \ac{nmt} models exhibit superior performance over \ac{smt} models in terms of \ac{cer}. Finally, Hamalainen et al. \citet{Hamalainen18} advocated for a combined approach involving \ac{smt}, \ac{nmt}, an edit-distance approach, and a rule-based finite-state transducer to leverage the strengths of each method.

\section{Normalization Approaches}
\label{se:approaches}
In this section we propose two new spelling normalization approaches based on \ac{llm}s, and describe two state-of-the-art approaches which will be used for comparison to evaluate our proposal. 

All approaches tackle spelling normalization from an \ac{mt} point of view. Given a source sentence $\x$, \ac{mt} aims to find the most likely translation $\hat{\y}$~\citep{Brown93}:

\begin{equation}
\hat{\y} = \argmax_{\y} Pr(\y \mid \x)
\label{eq:SMT}
\end{equation}

\subsection{mT5}
\label{se:mT5}
mT5 \citep{Xue21} is a multilingual variant of T5 \citep{Raffel20}, pre-trained on a new Common Crawl-based dataset covering 101 languages. We choose to use this \ac{llm} since it has been pre-trained without any supervised training and, thus, can be easily adapted to any downstream task by simply fine-tuning the model.

To that end, considering the original language as the source and its normalized version as the target language, our normalization approach consists in fine-tuning mT5 using the available training data.

\subsection{mBART}
mBART \citep{Liu20} is a multilingual encoder-decoder model primarily intended for translation tasks. This model was later extended for an extra set of 25 language tokens and, then, pre-trained on 50 languages \citep{Tang20}. We selected this \ac{llm} since it has already been pre-trained for \ac{mt} and covers a large variety of languages.

Similarly to the mT5 approach (see \cref{se:mT5}), this proposal consists in fine-tuning mBART using the available training data.

\subsection{Character-based SMT}
CBSMT computes \cref{eq:SMT} at a character level, using models that rely on a log-linear combination of different models \citep{Och02}: namely, phrase-based alignment models, reordering models and language models; among others~\citep{Zens02,Koehn03}.

We selected this approach since it still is part of the state of the art for some tasks \citep{Tang18,Hamalainen18,Domingo21b}. Considering the document's language as the source language and its normalized version as the target language, this approach splits words into characters and, then, applies a conventional \ac{smt} procedure.

\subsection{Character-based NMT}
\label{se:CBNMT}
This approach models \cref{eq:SMT} with a neural network using a Transformer architecture \citep{Vaswani17}. The source sentence is projected into a distributed representation at the encoding state. Then, the decoder generates, token by token, its most likely translation using a beam search method \citep{Sutskever14}. Model parameters are typically estimated jointly on large parallel corpora, via stochastic gradient descent \citep{Robbins51,Rumelhart86}. Finally, at decoding time, the system obtains the most likely translation by means of a beam search method.

Like the previous approach, this one tackles spelling normalization by considering the document's language as the source language and its normalized version as the target language. Words from the training documents are split into characters and, then, they are used for training an \ac{nmt} system.
\section{Experimental setup}
\label{se:exp}
In this section, we describe the conditions arranged in order to assess our proposal: evaluation metrics, corpora and normalization systems.

\subsection{Metrics}
We made use of the following well-known metrics in order to compare our different strategies:

\begin{description}
    \item[Character Error Rate (CER){\normalfont :}] number of character edit operations (insertion, substitution and deletion), normalized by the number of characters in the final translation.
    \item[Translation Error Rate (TER)] \citep{Snover06}: number of word edit operations (insertion, substitution, deletion and swapping), normalized by the number of words in the final translation.
    \item[BiLingual Evaluation Understudy (BLEU)]~\citep{Papineni02}: geometric average of the modified n-gram precision, multiplied by a brevity factor.
\end{description}

In order to ensure consistency with BLEU scores, we used \texttt{sacreBLEU} \citep{Post18}. Additionally, in order to determine whether two systems presented statistically significant differences, we applied approximate randomization tests \citep{Riezler05} with $10,000$ repetitions and using a $p$-value of $0.05$.

\begin{table}[!t]
    \resizebox{0.8\textwidth}{!}{\begin{minipage}{\textwidth}
    %\centering
    \begin{tabular}{c c c c c}
    \toprule
    &  & \textbf{Entremeses} & \multirow{2}{*}{\textbf{Quijote}} & \multirow{2}{*}{\textbf{Bohori\u{c}}} \\
    & & \textbf{y Comedias} & & \\
    \midrule
    \multirow{4}{*}{Train} & $|S|$ & 35.6K & 48.0K & 3.6K \\
    & $|T|$ & 250.0/244.0K & 436.0/428.0K & 61.2/61.0K \\
    & $|V|$ & 19.0/18.0K & 24.4/23.3K & 14.3/10.9K \\
    & $|W|$ & 52.4K & 97.5K & 33.0K \\
    \midrule
    \multirow{4}{*}{Dev.} & $|S|$ & 2.0K & 2.0K & 447 \\
    & $|T|$ & 13.7/13.6K & 19.0/18.0K & 7.1/7.1K \\
    & $|V|$ & 3.0/3.0K & 3.2/3.2K & 2.9/2.5K \\
    & $|W|$ & 1.9K & 4.5K & 3.8K \\
    \midrule
    \multirow{4}{*}{Test} & $|S|$ & 2.0K & 2.0K & 448 \\
    & $|T|$ & 15.0/13.3K & 18.0/18.0K & 7.3/7.3K \\
    & $|V|$ & 2.7/2.6K & 3.2/3.2K & 3.0/2.6K \\
    & $|W|$ & 3.3K & 3.8K & 3.8K \\
    \bottomrule
    \end{tabular}
    \end{minipage}}
    \caption{Corpora statistics. $|S|$ stands for number of sentences, $|T|$ for number of tokens, $|V|$ for size of the vocabulary and $|W|$ for the number of words whose spelling does not match modern standards. M denotes millions and K thousand.}
    \label{ta:corp}
\end{table}

\subsection{Corpora}
We made use of the following corpora in order to evaluate our proposal:

\begin{description}
    \item[Entremeses y Comedias] \citep{Jehle01}: A 17$^{\mathrm{th}}$ century Spanish collection of comedies by Miguel de Cervantes. It is composed of 16 plays, 8 of which have a very short length.
    \item[Quijote] \citep{Jehle01}: The 17$^{\mathrm{th}}$ century Spanish two-volumes novel by Miguel de Cervantes.
    \item[Bohori\u{c}] \citep{goo300k}: A collection of 18$^{\mathrm{th}}$ century Slovene texts written in the old Bohori\u{c} alphabet.
    %\item[Gaj] \citep{goo300k}: A collection of 19$^{\mathrm{th}}$ century Slovene texts written in the Gaj alphabet.
\end{description}

\cref{ta:corp} showcases the corpora statistics.

\begin{table*}[!ht]
    \centering
	\resizebox{0.8\textwidth}{!}{\begin{minipage}{\textwidth}
			\begin{tabular}{l c c c c c c c c c}
				\toprule
				\multirow{2}{*}{\textbf{System}} & \multicolumn{3}{c}{\textbf{Entremeses y Comedias}} & \multicolumn{3}{c}{\textbf{Quijote}} & \multicolumn{3}{c}{\textbf{Bohori\u{c}}} \\
				\cmidrule(lr){2-4}\cmidrule(lr){5-7}\cmidrule(lr){8-10}
				& CER [$\downarrow$] & TER [$\downarrow$] & BLEU [$\uparrow$] & CER [$\downarrow$] & TER [$\downarrow$] & BLEU [$\uparrow$] & CER [$\downarrow$] & TER [$\downarrow$] & BLEU [$\uparrow$] \\
				\midrule
				Baseline & $8.1$ & $28.0$ & $47.0$ & $7.9$ & $19.5$ & $59.4$ & $21.7$ & $49.0$ & $18.0$ \\
				CBSMT & $\mathbf{1.3\dagger}$ & $\mathbf{4.4}$ & $\mathbf{91.7\dagger}$ & $\mathbf{2.5\dagger}$ & $\mathbf{3.0\dagger}$ & $\mathbf{94.4\dagger}$ & $\mathbf{2.4}$ & $\mathbf{8.7}$ & $\mathbf{80.4}$ \\
				CBNMT & $\mathbf{1.4\dagger}$ & $6.1$ & $88.0\ddagger$ & $\mathbf{1.9\dagger}$ & $\mathbf{3.3\dagger}$ & $\mathbf{93.9\dagger}$ & $26.2$ & $30.6$ & $60.0$ \\
				\midrule
				mT5 & $4.8$ & $30.1\ddagger$ & $90.8\dagger$ & $\mathbf{1.7\dagger}$ & $\mathbf{2.5\dagger}$ & $\mathbf{95.5\dagger}$ & $48.9$ & $61.7$ & $39.0$ \\
				mBART & $5.8$ & $31.4\ddagger$ & $87.0\ddagger$ & $\mathbf{1.7\dagger}$ & $\mathbf{2.7\dagger}$ & $\mathbf{95.5\dagger}$ & $9.8$ & $39.1$ & $70.1$ \\
				\bottomrule
			\end{tabular}
	\end{minipage}}
	\caption{Experimental results. Baseline system corresponds to considering the original document as the document to which the spelling has been normalized to match modern standards. All results are significantly different between all systems except those denoted with $\dagger$ and $\ddagger$ (respectively). Best results are denoted in \textbf{bold}.}
	\label{ta:exp}
\end{table*}

\subsection{Systems}
\Ac{cbsmt} systems were trained with \emph{Moses} \citep{Koehn07}, following the standard procedure: a 5-gram language model{\textemdash}smoothed with the improved KneserNey method{\textemdash}was estimated using \emph{SRILM} \citep{Stolcke02}, and the weights of the log-linear model were optimized with MERT \citep{Och03a}.

\Ac{nmt} systems were built using \texttt{OpenNMT-py} \citep{Klein17}. We used $6$ layers; Transformer, with all dimensions set to $512$ except for the hidden Transformer feed-forward (which was set to $2048$); $8$ heads of Transformer self-attention; $2$ batches of words in a sequence to run the generator on in parallel; a dropout of $0.1$; Adam \citep{Kingma14}, using an Adam beta2 of $0.998$, a learning rate of $2$ and Noam learning rate decay with $8000$ warm up steps; label smoothing of $0.1$ \citep{Szegedy15}; beam search with a beam size of $6$; and joint \ac{bpe} applied to all corpora, using $32,000$ merge operations.

For fine-tuning both \ac{llm}s, we made use of the \texttt{HugginFace} library \citep{Wolf19}. We selected \emph{mt5-base} and \emph{mbart-large-50} as starting points.
\section{Results}
\label{se:res}
\cref{ta:exp} presents the results of our experimental session. As baseline, we assessed the spelling differences of the original documents with respect to their normalized version.

With two exceptions, all approaches successfully improved the baseline according to all metrics. Those exceptions use the \emph{Bohori\u{c}} corpus, and are most likely related to its small size and the nature of Slovene.

Both proposals have a similar behavior, except for \emph{Bohori\u{c}}, in which only the \emph{mBART} approach is able to yield improvements. Most likely, this is due to \emph{mBART} being already pre-trained for \ac{mt}, while \emph{mT5} has been pre-trained without any supervising training. Thus, the limited number of training samples from \emph{Bohori\u{c}} are not enough for correctly fine-tuning the model to perform spelling normalization.

Overall, the \ac{cbsmt} approached yielded the best performance in all cases and according to all metrics. The only exception is for \emph{El Quijote}, in which case all approaches presented similar performance with no statistically differences among their results. These results are coherent with other results reported in the literature \citep{Tang18,Hamalainen18,Domingo19c} and seem to indicate that \ac{smt} is still the most suitable technology for this task.

When comparing \emph{mBART} and \emph{mT5}, the former seems to be the most suitable for this task due to it having been pre-trained for \ac{mt}. Although their performance is very dependent to the data: they showcase no significant difference for \emph{El Quijote}, only a slight difference for \emph{Entremeses y Comedias} when measuring with CER (in which case \emph{mT5} has a better performance), and a significant difference for \emph{Bohori\u{c}}. However, when compared to training a Transformer model from scratch, none of the approaches have a clear advantage: the later performs significantly best for \emph{Entremeses y Comedias} and \emph{mBART} does it for \emph{Bohori\u{c}} (although it is worth noting that the \ac{cbnmt} approach does not improve the baseline in this case).

\subsection{In-depth comparison}
In this section, we study the behavior of each normalization approach when normalizing a sentence from each dataset. 
\begin{figure}[!ht]
	\resizebox{0.85\textwidth}{!}{\begin{minipage}{\textwidth}
			\begin{tabular}{r l}
				\textbf{Original:} &  {\textexclamdown}O mal logrado mo\c{c}o! Salid fuera; \\
				\textbf{Normalized:} & {\textexclamdown}O\textcolor{teal}{h} mal logrado mo\textcolor{teal}{z}o! Salid fuera; \\
				 & \\
				\textbf{CBSMT:} & {\textexclamdown}O\textcolor{teal}{h} mal logrado mo\textcolor{teal}{z}o! Sal\textcolor{red}{{\'i}}\textcolor{red}{\textvisiblespace} fuera; \\
				\textbf{CBNMT:} & {\textexclamdown}O\textcolor{teal}{h} mal logrado mo\textcolor{teal}{z}o! Sal\textcolor{red}{{\'i}}\textcolor{red}{\textvisiblespace} fuera; \\
				 & \\
				 \textbf{mT5:} & {\textexclamdown}O\textcolor{teal}{h} mal logrado mo\textcolor{teal}{z}o! Sali\textcolor{red}{\textvisiblespace} fuera ; \\
				 \textbf{mBART:} &  {\textexclamdown}O\textcolor{teal}{h} mal logrado mo\textcolor{teal}{z}o! Sal\textcolor{red}{{\'i}}\textcolor{red}{\textvisiblespace} fuera; \\
			\end{tabular}
	\end{minipage}}
	\caption{Example of modernizing a sentence from \emph{Entremeses y Comedias} with all the different approaches. \textvisiblespace\ denotes a character that has been removed as part of its normalization. Unnormalized characters that should have been normalized and wrongly normalized characters are denoted in \textcolor{red}{red}. Characters which were successfully normalized are denoted in \textcolor{teal}{teal}.}
	\label{fi:exEntremeses}
\end{figure}

\begin{figure*}[!ht]
	\resizebox{0.85\textwidth}{!}{\begin{minipage}{\textwidth}
			\begin{tabular}{r l}
				\textbf{Original:} & ``Para esso se yo vn buen remedio'', dixo el del Bosque; \\
				\textbf{Normalized:} & ``Para es\textcolor{teal}{\textvisiblespace}o s\textcolor{teal}{{\'e}} yo  \textcolor{teal}{u}n buen remedio'', di\textcolor{teal}{j}o el del Bosque; \\
				 & \\
				\textbf{CBSMT:} & ``Para es\textcolor{teal}{\textvisiblespace}o s\textcolor{teal}{{\'e}} yo  \textcolor{teal}{u}n buen remedio'', di\textcolor{teal}{j}o el del Bosque; \\
				\textbf{CBNMT:} & ``Para es\textcolor{teal}{\textvisiblespace}o s\textcolor{teal}{{\'e}} yo  \textcolor{teal}{u}n buen remedio'', di\textcolor{teal}{j}o el del Bosque; \\
				 & \\
				\textbf{mT5:} &  ``Para es\textcolor{teal}{\textvisiblespace}o s\textcolor{teal}{{\'e}} yo  \textcolor{teal}{u}n buen remedio'', di\textcolor{teal}{j}o el \textcolor{red}{\textvisiblespace}\textcolor{red}{\textvisiblespace}\textcolor{red}{\textvisiblespace} \textcolor{red}{\textvisiblespace}\textcolor{red}{\textvisiblespace}\textcolor{red}{\textvisiblespace}\textcolor{red}{\textvisiblespace}\textcolor{red}{\textvisiblespace}\textcolor{red}{\textvisiblespace}\textcolor{red}{\textvisiblespace} \\
				\textbf{mBART:} &  ``Para es\textcolor{teal}{\textvisiblespace}o s\textcolor{teal}{{\'e}} yo  \textcolor{teal}{u}n buen remedio'', di\textcolor{teal}{j}o el \textcolor{red}{\textvisiblespace}\textcolor{red}{\textvisiblespace}\textcolor{red}{\textvisiblespace} \textcolor{red}{\textvisiblespace}\textcolor{red}{\textvisiblespace}\textcolor{red}{\textvisiblespace}\textcolor{red}{\textvisiblespace}\textcolor{red}{\textvisiblespace}\textcolor{red}{\textvisiblespace}\textcolor{red}{\textvisiblespace} \\
			\end{tabular}
	\end{minipage}}
	\caption{Example of modernizing a sentence from \emph{El Quijote} with all the different approaches. \textvisiblespace\ denotes a character that has been removed as part of its normalization. Unnormalized characters that should have been normalized and wrongly normalized characters are denoted in \textcolor{red}{red}. Characters which were successfully normalized are denoted in \textcolor{teal}{teal}.}
	\label{fi:exQuijote}
\end{figure*}

\cref{fi:exEntremeses} shows an example from \emph{Entremeses y Comedias}. In this case, the normalization only affects two characters. All approaches were able to successfully normalize those characters. However, they introduced new errors when normalizing the word \emph{Salid}: CBSMT, CBNMT and mBART both add an accent to the letter \emph{i} and remove the letter \emph{d} (which is the result of changing the tense of the verb from imperative to past simple), while mT5 removes the letter \emph{d} (which is plainly an orthographic mistake).

In the example from \emph{El Quijote} (see \cref{fi:exQuijote}), there are four characters that need to be normalized. In this case, CBSMT and CBNMT successfully normalize the sentence. mT5 and mBART, however, while they are able to normalize those four characters, they end up missing the final part of the sentence.

\begin{figure*}[!ht]
	\resizebox{0.85\textwidth}{!}{\begin{minipage}{\textwidth}
			\begin{tabular}{r l}
				\textbf{Original:} & vadljajo ali l{\'o}fajo, de bi sv{\'e}dili, kdo jim je kriv te nefrezhe. \\
				\textbf{Normalized:} & vadljajo ali l\textcolor{teal}{os}ajo, d\textcolor{teal}{a} bi \textcolor{teal}{iz}v\textcolor{teal}{e}d\textcolor{teal}{e}li, kdo jim je kriv te ne\textcolor{teal}{s}re\textcolor{teal}{\u{c}\textvisiblespace}e. \\
				 & \\
				\textbf{CBSMT:} & vadljajo ali l\textcolor{teal}{os}ajo, d\textcolor{teal}{a} bi \textcolor{teal}{iz}v\textcolor{teal}{e}d\textcolor{teal}{e}li, kdo jim je kriv te ne\textcolor{teal}{s}re\textcolor{teal}{\u{c}\textvisiblespace}e. \\
				\textbf{CBNMT:} & vadljajo ali l\textcolor{teal}{os}ajo, d\textcolor{teal}{a} bi \textcolor{red}{\textvisiblespace}\textcolor{teal}{z}v\textcolor{teal}{e}d\textcolor{red}{i}li, kdo jim je kriv te ne\textcolor{teal}{s}re\textcolor{teal}{\u{c}\textvisiblespace}e. \\
				 & \\
				 \textbf{mT5:} & vadljajo ali l\textcolor{red}{a\u{z}e}jo, da bi \textcolor{red}{\textvisiblespace}\textcolor{red}{\textvisiblespace}v\textcolor{teal}{e}d\textcolor{teal}{e}li, kdo jim je kriv te \textcolor{red}{\textvisiblespace}\textcolor{red}{\textvisiblespace}\textcolor{red}{\textvisiblespace}\textcolor{red}{\textvisiblespace}\textcolor{red}{\textvisiblespace}\textcolor{red}{\textvisiblespace}\textcolor{red}{\textvisiblespace}\textcolor{red}{\textvisiblespace}\textcolor{red}{\textvisiblespace} \\
				 \textbf{mBART:} & vadljajo ali l\textcolor{teal}{os}ajo, d\textcolor{teal}{a} bi \textcolor{red}{\textvisiblespace}\textcolor{red}{\textvisiblespace}v\textcolor{teal}{e}d\textcolor{teal}{e}li, kdo jim je kriv te ne\textcolor{teal}{s}re\textcolor{teal}{\u{c}\textvisiblespace}e. \\
			\end{tabular}
	\end{minipage}}
	\caption{Example of modernizing a sentence from \emph{Bohori\u{c}} with all the different approaches. \textvisiblespace\ denotes a character that has been removed as part of its normalization. Unnormalized characters that should have been normalized and wrongly normalized characters are denoted in \textcolor{red}{red}. Characters which were successfully normalized are denoted in \textcolor{teal}{teal}.}
	\label{fi:exBohoric}
\end{figure*}

Finally,  \cref{fi:exBohoric} shows an example from \emph{Bohori\u{c}}. In this example, ten characters from four words need to be normalized. As with the previous dataset, the CBSMT approach successfully normalizes the sentence. 

The CBNMT approach successfully normalizes three of the words and makes a mistake with two of the characters from one word: one character{\textemdash}which did not exist in the original word{\textemdash}is still missing, and another one is left unnormalized.

mBART is also able to successfully normalize three of the words, and makes a mistake with two of the characters from one word: the same missing character than in the previous approach, and it also removes the character \emph{s} instead of normalizing it as \emph{z}.

Finally, mT5 has the worst performance: it makes the same mistakes as mBART when normalizing the word \emph{sv{\'e}dili}, makes several new mistakes when normalizing the word \emph{l{\'o}fajo} (failing to normalize the two characters that needed to be normalized, and changing one which did not need to be changed) and the word \emph{nefrezhe.} is missing from the hypothesis.
\section{Conclusions and future work}
\label{se:conc}
In this work we proposed two spelling normalization approaches based on \ac{llm}s: one trained without a supervised training and another which had been trained for \ac{mt}. We assessed our proposals on several datasets from different time periods and languages, and compared their performance against two other approaches from the literature.

Results showed that \emph{mT5} is harder to adapt to spelling normalization, most likely due to the scarce availability of parallel training data when working with historical documents~\citep{Bollmann16}. The \emph{mBART} approach had a better performance and successfully improved all baselines. However, it was not able to improve results yielded by other performance: it reached the same results for one dataset, and slightly worse results for the other two (in one of which presented the second best results).

Overall, \ac{cbsmt} achieved the best results, leading to the conclusion that it is still the most suitable technology for this task.

As a future work, we would like to explore the use of other \ac{llm}s. Additionally, we would like to try to combine the different datasets and explore the use of old documents (for which no normalization is available) to do a general fine-tuning of the \ac{llm}, instead of fine-tuning the model for each historical period.
%\section*{Acknowledgments}

% \section*{Limitations}
% In this work we have only tested two different \ac{llm}s. Thus, while we concluded that \ac{smt} seems to still be the most suitable technology for this task, we cannot infer that \ac{llm}s are not suitable without extending the study to more models.

% While we selected \emph{mT5} due to it being easily adapted to any downstream task, we observed that it is heavily affected by the limited availability of training data in this field of work. However, as a first approach, in this work we have only study fine-tuning the model for each historical context. In a future work we need to study weather combining the datasets and doing a single fine-tuning for the task{textemdash}instead of per language and historical time frame{\textemdash}benefits the model.

\bibliographystyle{apalike}
\bibliography{snor}

\end{document}